\documentclass[final]{article}

\usepackage[final]{neurips_2025}

\usepackage[utf8]{inputenc} %
\usepackage[T1]{fontenc}    %
\usepackage{hyperref}       %
\usepackage{url}            %
\usepackage{booktabs}       %
\usepackage{amsfonts}       %
\usepackage{nicefrac}       %
\usepackage{microtype}      %
\usepackage[table]{xcolor}  %
\usepackage[english]{babel}
\usepackage{blindtext}
\usepackage{xspace}
\usepackage{amsmath}
\usepackage[normalem]{ulem}
\usepackage{caption}
\usepackage{graphicx}
\usepackage{booktabs}
\usepackage{arydshln} %
\usepackage{xurl}
\usepackage{wrapfig}
\usepackage[toc,title]{appendix}
\usepackage{afterpage}
\usepackage{tikz}
\usepackage{enumitem}
\usepackage{stfloats}

\usepackage{comment}

\usepackage{multirow}

\usepackage{xspace}
\usepackage{mfirstuc} %
\usepackage{longtable}
\usepackage{listings}
\usepackage[most]{tcolorbox} %
\usepackage{graphicx} %
\usepackage{multicol} 
\usepackage{subcaption}

\usepackage{siunitx}

\newcommand{\sys}{\textit{LLMVisor}\xspace}
\newcommand{\kv}{\text{KV cache}\xspace}

  {\begin{list}{\labelitemi}{\itemsep2pt \topsep2pt \parsep0.00in
  \partopsep=0pt \leftmargin1.2em}}%
  {\end{list}}

\let\latexusecounter=\usecounter
\def\compactsortof{\itemsep=2pt \topsep=2pt \parsep=0.00in \partopsep=0pt
\leftmargin=1.2em}

\newcommand{\BULLET}{\vspace{+.00in} \noindent $\bullet$ \hspace{+.00in}}

\usepackage{enumitem}
\setlist[itemize]{nosep, itemsep=0pt, topsep=0pt, leftmargin=1em, labelwidth=*, align=left}
\setlist[enumerate]{nosep, itemsep=0pt, topsep=0pt, leftmargin=1em, labelwidth=*, align=left}

\usepackage{hyperref}

\title{LLMVisor: A Real-Time Latency Attribution Model for Multi-Tenant LLM Serving}

\author{%
\begin{tabular}{c}
Shuowei Jin\textsuperscript{1}\thanks{Equal contribution.}\quad
Xueshen Liu\textsuperscript{1}\footnotemark[1]\quad
Jiaxin Shan\textsuperscript{2}\quad
Le Xu\textsuperscript{2}\\
Tieying Zhang\textsuperscript{2}\quad
Liguang Xie\textsuperscript{2}\quad
Z.~Morley Mao\textsuperscript{1}\\[1ex]
\textsuperscript{1}University of Michigan\quad
\textsuperscript{2}ByteDance
\end{tabular}
}
\vspace{-0.2in}

\begin{document}

\maketitle

\begin{abstract}
As LLM inference shifts to multi-tenant GPU clusters, co-batching improves throughput but obscures per-tenant usage and limits control. Enabling fractional sharing of the inference engine requires a real-time, per-request attribution primitive that is accurate and light enough to run inside the scheduling loop. We present \sys, a roofline-guided \emph{latency attribution} model that captures the memory-bound and compute-bound phases via a concise piecewise-linear form over features proportional to FLOPs and memory I/O traffic. \sys decomposes batch latency into additive, per-request shares and runs efficiently at $\mu$s-scale. We evaluate \sys across Llama3.1-8B and Qwen2.5-14B/32B on A100/H100 under varying tensor parallelism and workload mixes. Compared to a token-count baseline, \sys attains near-perfect $R^2$ and reduces relative error by up to $2.5\times/3.3\times$ (p90/p99) for prefill and $3.5\times/4.4\times$ for decode, despite batching variability and sequence divergence. 
\end{abstract}

\section{Introduction}

Large language models (LLMs) now power search, coding assistants, and conversational systems at scale~\cite{bommasani2021opportunities,jin2025search,hurst2024gpt,jin2025plato,wu2025hetermoe,wu2025rlboost,zhao2024eagle}. To keep accelerators busy, providers co-batch heterogeneous requests via shared public endpoints~\cite{openai-api,claude-api,team2025aibrix}. While batching boosts throughput, it leaves tenants with opaque, provider-tuned scheduling and little leverage to allocate or account for their own usage. The common alternative—renting dedicated GPU servers—squanders parallel hardware on interactive, small-batch workloads; for example, GPT-J~6B on A100 achieves only 0.4\% SM utilization at batch size~1 and remains low even for 1{,}024-token sequences~\cite{jin2023s}.

We argue for virtualizing the \emph{inference engine}: tenants reserve fractional shares of GPU time and submit requests into shared batches, akin to CPU isolation via VMs, cgroups, and containers~\cite{barham2003xen,menage2008cgroups,merkel2014docker}, and to credit/burst offerings (e.g., AWS \texttt{t3.*})~\cite{aws_t3}. Hardware partitioning (e.g., NVIDIA MIG) provides strong isolation but is inflexible for LLM memory footprints~\cite{nvidia_mig_user_guide}. Realizing a software path to fractional sharing hinges on a missing primitive: a \emph{real-time, per-request latency attribution model} accurate enough to guide scheduling and light enough to run in the critical path without harming batching efficiency. Such attribution directly enables SLO-aware admission (curbing tails by rejecting or deferring requests whose shares would exceed budgets), transparent post-hoc accounting, and fast “what-if’’ planning (e.g., alternative batch sizes or mixes) while preserving the benefits of co-batching.

Designing this primitive is difficult for several reasons. First, co-batching couples requests: their costs are not trivially separable. Second, LLM inference has phase-specific bottlenecks, compute-bound \emph{prefill} versus memory-bound \emph{decode}, that require different treatments. Third, costs scale nonlinearly with sequence length (e.g., quadratic self-attention) and depend on context length (KV-cache loads) and batch size. Fourth, throughput shifts with model architecture, tensor-parallel layout, and hardware. Finally, the scheduler operates at millisecond granularity, so attribution must run at $\mu$s scale to be viable. Black-box ML predictors can fit latency trends but lack additivity and are too slow for the loop~\cite{yuan2024llm}; naive token-count proxies~\cite{sheng2024vtc} ignore context and batching effects, yielding poor high-percentile accuracy.

We present \sys, a roofline-guided \emph{latency attribution} model for multi-tenant LLM serving. \sys captures prefill and decode with a concise piecewise-linear form over features proportional to FLOPs and memory I/O traffic—explicitly encoding quadratic self-attention, context-length–driven KV-cache traffic, and batch-size effects—so end-to-end batch latency decomposes into additive, per-request shares. The resulting structure supports online $\mu$s-scale “what-if’’ queries for admission control and budgeting while preserving batching efficiency. A lightweight prototype integrated into vLLM runs in the scheduling path with negligible overhead and achieves low error across diverse models and GPUs.

Our contributions are threefold:

\BULLET \textbf{Inference-aware modeling:} An interpretable, roofline-guided, piecewise-linear formulation that embeds LLM execution mechanics: compute-bound prefill vs.\ memory-bound decode, quadratic self-attention, KV-cache traffic via context length, and batch-size utilization effects. The model is additive by design, yielding closed-form per-request latency shares suitable for multi-tenant control.

\BULLET  \textbf{Efficiency:} Microsecond-level runtime per scheduling step (100x faster than common ML predictors), enabling deployment \emph{inside} the scheduler without disrupting batching.

\BULLET  \textbf{Accuracy and generality:} Across Llama3.1-8B and Qwen2.5-14B/32B on A100/H100 with varying tensor parallelism and workload mixes, \sys attains near-perfect $R^2$ and reduces relative error by up to $2.5\times/3.3\times$ (p90/p99) for prefill and $3.5\times/4.4\times$ for decode.

\section{\sys Design}
\label{sec:design}

\begin{figure}[t] %
    \centering
    \includegraphics[width=0.8\linewidth]{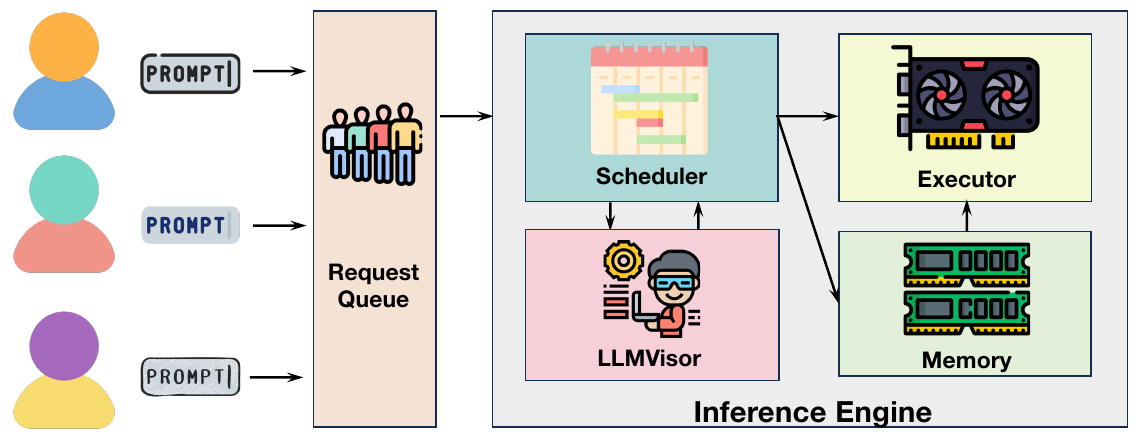} %
    \caption{A typical workflow in modern large language model (LLM) inference engines can be summarized as follows. Upon arrival, each request is enqueued in a request buffer. At each inference step, the scheduler selects requests from the queue for execution. \sys operates during the scheduling stage by providing GPU time attribution, enabling the scheduler to enforce multi-tenant fairness according to reserved GPU time quotas. Once scheduled, the inference engine dispatches the requests to the GPU, where the corresponding \kv and model weights are loaded to perform the inference computation.
}
    \label{fig:design}
    \vspace{-0.2in}
\end{figure}

\textbf{Design requirements.}
The state-of-the-art inference engine workflow is illustrated in Fig.~\ref{fig:design}. \sys assists the scheduler in ensuring that each tenant receives its reserved share of GPU time. To be practical in high-frequency scheduling loops (e.g., every decode step), \sys{} must combine three properties: it must be \textbf{\emph{accurate}}, providing reliable predictions of end-to-end batch latency; it must support \textbf{\emph{attribution fidelity}}, decomposing that latency into per-request and per-tenant contributions; and it must be \textbf{\emph{efficient},} introducing negligible overhead so that scheduling can operate at each step\footnote{Step is the minimal execution unit in LLM inference. During decoding, one step corresponds to all active requests in the batch generating a single token. During prefill, one step corresponds to all requests in the batch completing their entire prompt processing.
}. \sys{} is designed to satisfy all three simultaneously.

\textbf{Problem formulation.}
Formally, let $\mathcal{B}=\{1,\dots,|\mathcal{B}|\}$ denote a batch of requests and $u(i)$ the tenant of request $i$. We seek a model $f$ that (i) predicts the end-to-end latency of the batch, $f(\mathcal{B}) \approx T_{\mathcal{B}}$, and (ii) assigns each request a nonnegative share $f(i \mid \mathcal{B}) \approx t_i$ such that $\sum_{i\in\mathcal{B}} f(i \mid \mathcal{B}) = f(\mathcal{B})$. These per-request attributions capture each request’s latency contribution and aggregate naturally to per-tenant usage $L_u = \sum_{i:\,u(i)=u} f(i \mid \mathcal{B})$, supporting fairness, admission control, and billing.

\textbf{Related work.}
Prior studies on LLM latency prediction follow two main approaches. The first uses ML predictors~\cite{imai2024predicting}, which can be accurate but cannot attribute latency to individual requests, limiting their usefulness for multi-tenant scheduling. The second relies on analytical, roofline-style modeling~\cite{yuan2024llm}, classifying inference as compute- or memory-bound and estimating latency from FLOPs and memory traffic relative to GPU peaks. While interpretable, these models are often inaccurate in practice due to modeling assumptions and time-varying GPU throughput, and they do not provide lightweight, per-request attribution in real time. VTC~\cite{sheng2024vtc} attributes usage by token counts, which is imprecise because compute and I/O vary with token position~\cite{jin2024compute}.

\subsection{Piecewise roofline-guided linear latency model}

\sys{} builds on the intuition of roofline analysis but adapts it for higher accuracy and real-time use. Rather than hand-computing FLOPs and memory bytes from model architecture and dividing by peak GPU rates, we construct a small set of request-level features proportional to FLOPs and I/O traffic and fit their coefficients using short warm-up profiling runs. This preserves interpretability while improving accuracy and keeping runtime cost negligible.

Let $p_i$ denote the number of tokens \emph{to process} for request $i$ in the current step (all prompt tokens during prefill; $p_i=1$ during decode), let $c_i$ denote the \emph{context} tokens that must be attended and whose \kv must be loaded (zero in prefill; prior prompt length in decode), and let $\mathcal{B}$ denote the batch size. As $\sum_i p_i$ increases, the feed-forward network (FFN) transitions from memory-bound to compute-bound. Thus, we model both prefill and decode latency using a two-segment linear form with different coefficient sets $(\beta_{\star}, \{a_1, a_2, a_3, a_4\}_{\star})$.

\begin{equation}
T_{\mathcal{B}} =
\beta + a_1 \!\sum_i p_i + a_2 \!\sum_i c_i + a_3 \!\sum_i p_i^2 + a_4 \!\sum_i |\mathcal{B}|
\label{eq:split}
\end{equation}

Here, $\beta$ captures fixed per-batch costs (e.g., model-weight I/O); $\sum_i p_i$ reflects MLP and causal-attention compute; $\sum_i c_i$ accounts for KV-cache memory traffic; $\sum_i p_i^2$ models quadratic self-attention costs; and $\sum_i |\mathcal{B}|$ captures utilization gain. We fit the coefficients via ordinary least squares on profiling data, yielding four sets of parameters $(\beta_{\star}, \{a_1, a_2, a_3, a_4\}_{\star})$ in total—two for prefill and two for decode.

Because $T_{\mathcal{B}}$ is linear in sums of per-request terms, \sys{} naturally admits \textbf{closed-form}, additive per-request attributions via Eq.~\ref{eq:split}. yielding each request’s estimated latency contribution in the critical path.

Additionally, in terms of \textbf{efficiency}, Eq.~\ref{eq:split} is over 100x faster than ML predictors such as Random Forest. In our measurements, \sys{} operates at the microsecond level per request, whereas more complex ML models like Random Forest require milliseconds for inference. This efficiency makes \sys{} well-suited for integration into LLM inference engine scheduling, which runs at millisecond granularity.

\section{Evaluation}

In this section, we evaluate \sys across a diverse range of real-world LLM serving configurations and compare its predictions against ground truth measurements.

\subsection{Experimental Setup}
We evaluate \sys inside vLLM (v0.7.3)~\citep{kwon2023vllm} using its internal profiler for per-step ground truth, spanning two server classes (H100 SXM and A100 SXM) with multiple tensor parallel sizes, three open-source models (Llama3.1-8B~\citep{meta2024llama31blog}, Qwen2.5-14B/32B~\citep{qwen2.5}), and workloads with 128–2048 concurrent requests and heterogeneous sequences that push total tokens to ninety percent of the engine KV-cache capacity ($\mathcal{C}$). We cover both prefill and decode, report coefficient of determination ($r^2$) and relative error at the 90th and 99th percentiles, and compare against VTC~\citep{sheng2024vtc}, a token based proxy that does not model batch size or context length. Further details of the evaluation setup are provided in Sec.~\ref{sec:appendix}.

\begin{table}[t]
\small
\centering
\begin{tabular}{l|ccc|ccc|ccc} 
\hline 
Model          & \multicolumn{3}{c|}{Llama3.1-8B} & \multicolumn{3}{c|}{Qwen2.5-14B} & \multicolumn{3}{c}{Qwen2.5-32B} \\ 
\hline
Parallelism            & A1    & H1    & H4    & H1    & H2    & H4    & A2    & H2    & H4          \\ 
\hline 
VTC $R^2$              & 0.998 & 0.998 & 0.995 & 1.000 & 0.997 & 0.998 & 1.000 & 1.000 & 1.000 \\ 
VTC $\text{Err}_{p90}$ & 0.08  & 0.09  & 0.04  & 0.03  & \color{blue}{0.14}  & 0.08  & 0.02  & 0.03  & 0.02 \\ 
VTC $\text{Err}_{p99}$ & 0.50  & 0.79  & 0.16  & 0.09  & \color{blue}{1.24}  & 0.57  & 0.10  & 0.07  & 0.11 \\ 

\hline \hline
\sys $R^2$              & 1.000 & 0.999 & 1.000 & 1.000 & 1.000 & 1.000 & 1.000 & 1.000 & 1.000 \\ 
\sys $\text{Err}_{p90}$ & 0.03  & \color{blue}{0.10}  & 0.01  & 0.02  & 0.02  & 0.01  & 0.01  & 0.02  & 0.03  \\ 
\sys $\text{Err}_{p99}$ & 0.14  & \color{blue}{0.92}  & 0.08  & 0.08  & 0.08  & 0.13  & 0.05  & 0.07  & 0.08  \\ 
\hline
\end{tabular}
\caption{Prefill step latency modeling accuracy of VTC and \sys.}
\label{tab:prefill_fit}
\end{table}

\begin{table}[t]
\small
\centering
\begin{tabular}{l|ccc|ccc|ccc} 
\hline 
Model          & \multicolumn{3}{c|}{Llama3.1-8B} & \multicolumn{3}{c|}{Qwen2.5-14B} & \multicolumn{3}{c}{Qwen2.5-32B} \\ 
\hline
Parallelism            & A1    & H1    & H2    & H1    & H2    & H4    & A2    & H2    & H4          \\ 
\hline 
VTC $R^2$              & 0.947 & 0.958 & 0.778 & 0.991 & 0.948 & 0.938 & 0.972 & 0.991 & 0.992 \\ 
VTC $\text{Err}_{p90}$ & 0.23  & 0.24  & 0.42  & 0.12  & 0.24  & 0.27  & 0.15  & 0.12  & 0.11 \\ 
VTC $\text{Err}_{p99}$ & 0.50  & 0.54  & 0.88  & 0.18  & 0.45  & 0.63  & 0.29  & 0.20  & 0.31 \\ 

\hline \hline
\sys $R^2$              & 0.974 & 0.997 & 0.995 & 0.998 & 0.997 & 0.996 & 0.979 & 0.997 & 0.999 \\ 
\sys $\text{Err}_{p90}$ & 0.10  & 0.04  & 0.05  & 0.04  & 0.04  & 0.07  & 0.11  & 0.06  & 0.03  \\ 
\sys $\text{Err}_{p99}$ & 0.16  & 0.08  & 0.09  & 0.06  & 0.07  & 0.10  & 0.16  & 0.10  & 0.05  \\ 
\hline
\end{tabular}
\caption{Decode step latency modeling accuracy of VTC and \sys.}
\vspace{-0.2in}
\label{tab:decode_fit}
\end{table}

\subsection{Prefill Latency Modeling}
The performance of \sys and the VTC baseline on the prefill latency modeling task is presented in \autoref{tab:prefill_fit}. Across all evaluated models and hardware configurations, \sys consistently outperforms VTC in modeling prefill latency. 

While VTC achieves reasonably high $R^2$ values ($>0.995$ in most cases), it exhibits much larger relative errors, particularly at p99. Averaged across all configurations (excluding the two blue-marked \textcolor{blue}{outliers}), VTC’s relative error is $0.05$ at p90 and $0.30$ at p99, while \sys reduces these to $0.02$ and $0.09$, respectively. This corresponds to an average improvement of roughly $2.5\times$ at p90 and over $3.3\times$ at p99. The high $R^2$ values for both methods indicate that VTC’s linear token-based model can capture coarse latency trends in prefill phase. However, by taking into account the square of input tokens, which reflects the self-attention computational complexity, \sys is far more reliable for predicting prefill latency in multi-tenant serving environments.

\subsection{Decode Latency Modeling}
The performance of \sys and the VTC baseline on the decode latency modeling task is presented in \autoref{tab:decode_fit}. Compared to the prefill phase, decode latency is substantially more challenging to predict. This difficulty is evident in VTC’s results: while it achieves moderately strong correlation in most cases ($R^2 > 0.9$), it completely breaks down in others. For example, Llama3.1-8B on H2 yields only $R^2=0.778$, indicating that VTC is not even able to capture the overall latency trend. At the same time, its relative errors remain high across the board, averaging 0.21 at p90 and 0.44 at p99, and reaching as high as 0.88 at p99. These results show that a simple linear token-based model is fundamentally inadequate for decode latency, as it ignores the effects of context length and batch size. 

In contrast, \sys consistently maintains near-perfect fits, with $R^2>0.97$ in all cases and typically above 0.995. More importantly, \sys reduces relative error dramatically: average p90 error drops from 0.21 (VTC) to 0.06, and average p99 error from 0.44 to 0.10, corresponding to $3.5\times$ and $4.4\times$ improvements, respectively. Even in the more challenging decode phase, where latency is influenced by batching variability and sequence divergence, \sys reliably captures both the overall trend and high-percentile behavior.

Overall, \sys significantly outperforms linear token-based modeling in both prefill and decode phases, with up to 3.5x and 4.4x improvement on p90 and p99 relative error, respectively. It consistently predicts latency accurately and reliably, which will benefit the scheduling in multi-tenant serving environments.

\bibliography{reference}
\bibliographystyle{plain}

\clearpage

\appendix
\section{Evaluation Setup Details}
\label{sec:appendix}
Our evaluation framework is built on vLLM (v0.7.3)~\citep{kwon2023vllm}, using its internal profiler to capture ground truth latency for each processing step. To ensure a comprehensive assessment, we systematically vary the hardware, models, and workloads.

\textbf{Hardware} We conduct experiments on two distinct server types with five different tensor parallelism settings to cover various deployment scenarios:
a. A server with 4 NVIDIA H100 SXM 80GB GPUs, a 104-core vCPU, and 1800GiB of memory. Tensor parallel size 1, 2, and 4, denoted as \textbf{H1}, \textbf{H2}, \textbf{H4}. \quad
b. A server with 8 NVIDIA A100 SXM 80GB GPUs, a 240-core vCPU, and 1800GiB of memory. Tensor parallel size 1 and 2, denoted as \textbf{A1}, \textbf{A2}.

\textbf{Models} We test \sys on three popular open-source LLMs of varying sizes to evaluate its generalizability: Llama3.1-8B~\citep{meta2024llama31blog}, Qwen2.5-14B~\citep{qwen2.5}, and Qwen2.5-32B~\citep{qwen2.5}.

\textbf{Workloads} To simulate realistic, multi-tenant serving environments, we generate workloads with varying numbers of concurrent requests, from 128 to 2048. To test performance across different context lengths, we increase the total number of tokens from 4096 to 90\% of the engine's maximum KV cache capacity ($\mathcal{C}$). These tokens are then randomly distributed among requests with high variance to ensure a heterogeneous mix of request lengths. Our evaluation covers both the prefill and decode phases of inference.

\textbf{Metrics} We assess the accuracy of resource modeling methods via two metrics: 
a. Coefficient of Determination ($r^2$): Quantifies the linear correlation between the predicted and ground truth latencies. A higher $r^2$ indicates a better fit. \quad
b. Relative Error: Measures the percentage difference between predicted and actual latency, reported at the 90th (p90) and 99th (p99) percentiles to evaluate performance.

\textbf{Baseline} We benchmark \sys against VTC~\citep{sheng2024vtc}, a baseline model that estimates request latency based on a linear model of token number. Unlike \sys, VTC does not account for critical system-level factors such as batch size and context length.

\end{document}